# Memetic Search in Differential Evolution Algorithm

Sandeep Kumar
Jagannath University
Chaksu, Jaipur, Rajasthan,
India - 303901

Vivek Kumar Sharma, Ph.D
Jagannath University
Chaksu, Jaipur, Rajasthan,
India - 303901

Rajani Kumari
Jagannath University
Chaksu, Jaipur, Rajasthan,
India - 303901

## ABSTRACT
Differential Evolution (DE) is a renowned optimization stratagem that can easily solve nonlinear and comprehensive problems. DE is a well known and uncomplicated population based probabilistic approach for comprehensive optimization. It has apparently outperformed a number of Evolutionary Algorithms and further search heuristics in the vein of Particle Swarm Optimization at what time of testing over both yardstick and actual world problems. Nevertheless, DE, like other probabilistic optimization algorithms, from time to time exhibits precipitate convergence and stagnates at suboptimal position. In order to stay away from stagnation behavior while maintaining an excellent convergence speed, an innovative search strategy is introduced, named memetic search in DE. In the planned strategy, positions update equation customized as per a memetic search stratagem. In this strategy a better solution participates more times in the position modernize procedure. The position update equation is inspired from the memetic search in artificial bee colony algorithm. The proposed strategy is named as Memetic Search in Differential Evolution (MSDE). To prove efficiency and efficacy of MSDE, it is tested over 8 benchmark optimization problems and three real world optimization problems. A comparative analysis has also been carried out among proposed MSDE and original DE. Results show that the anticipated algorithm go one better than the basic DE and its recent deviations in a good number of the experiments.

## General Terms
Computer Science, Nature Inspired Algorithms, Meta-heuristics

## Keywords
Differential Evolution, Swarm intelligence, Evolutionary computation, Memetic algorithm

## 1. INTRODUCTION
Population-based optimization algorithms find near-optimal solutions to the easier said than done optimization problems by inspiration from nature or natural entities. A widespread characteristic of each and every one population-based algorithm is that the population consisting of potential solutions to the problem is tailored by applying some operators on the solutions depending on the information of their fitness. Hence, the population is moved towards better solution areas of the search space. Two essential classes of population-based optimization algorithms are evolutionary algorithms [1] and swarm intelligence-based algorithms [2]. Although Genetic Algorithm (GA) [3], Genetic Programming (GP) [4], Evolution Strategy (ES) [5] and Evolutionary Programming (EP) [6] are popular evolutionary algorithms, DE comes underneath the class of Evolutionary algorithms. Among an assortment of EAs, differential evolution (DE), which characterized by the diverse mutation operator and antagonism strategy from the other EAs, has shown immense promise in many numerical benchmark problems as well as real-world optimization problems. Differential evolution (DE) is a stochastic, population-based search strategy anticipated by Storn and Price [9] in 1995. While DE shares similarities with other evolutionary algorithms (EA), it differs considerably in the sense that distance and direction information from the current population is used to steer the search progression. In DE algorithm, all solutions have an equal opportunity of being preferred as parents, i.e. selection does not depend on their fitness values. In DE, each new solution fashioned competes with its parent and the superior one wins the contest. DE intensification and diversification capabilities depend on the two processes, that is to say mutation process and crossover process. In these two processes, intensification and diversification are evenhanded using the fine tuning of two parameters that is to say scale factor 'F' and crossover probability 'CR'. In DE the child vector is generated by applying the mutation and crossover operation. In mutation operation, a trial vector is generated with the help of the objective vector and two erratically preferred individuals. The perturbation in objective vector depends on F and the difference between the randomly selected individuals. Further, a crossover operation is applied between the objective vector and parent vector for generating the child vector using crossover probability (CR). So, it is clear that the discrepancy in the child vector from the parent vector depends on the values of F and CR. This discrepancy is measured as a step size for the candidate solution/parent vector. Large step size may results in skipping of actual solutions that is due to large value of CR and F while if these values are low then the step size will be small and performance degrades.

The hybridization of some local search techniques in DE may diminish the possibility of skipping factual solution. In hybridized search algorithms, the global search capability of the main algorithm explore the search space, trying to identify the most promising search space regions while the local search part scrutinizes the surroundings of some initial solution, exploiting it in this way. Therefore, steps sizes play an important role in exploiting the identified region and these step sizes can be controlled by incorporating a local search algorithm with the global search algorithm.

Differential evolution (DE) has come into sight as one of the high-speed, robust, and well-organized global search heuristics of in progress significance. Over and over again real world provides some very complex optimization problems that cannot be easily dealt with existing classical mathematical optimization methods. If the user is not very susceptible about the exact solution of the problem in hand then nature-inspired algorithms may be used to solve these kinds of problems. It is shown in this paper that the nature-inspired algorithms have been gaining to a great extent of recognition now a day due to the fact that numerous real-world optimization problems have turn out to be progressively more large, multifarious and self-motivated. The size and complication of the problems at the present time necessitate the development of methods and solutions whose efficiency is considered by their ability to find up to standard results within a levelheaded amount of time, rather than an aptitude to guarantee the optimal solution. The paper also includes a concise appraisal of well-organized and newly developed nature-inspired algorithm, that is to say Differential Evolution. In addition, to get better effectiveness, correctness and trustworthiness, the considered algorithm is analyzed, research gaps are identified. DE has apparently outperformed a number of Evolutionary Algorithms (EAs) and other search heuristics in the vein of the Particle Swarm Optimization (PSO) at what time tested over together benchmark and real world problems. The scale factor (F) and crossover probability (CR) are the two major parameters which controls the performance of DE in its mutation and crossover processes by maintain the balance between intensification and diversification in search space. Literature suggests that due to large step sizes, DE is somewhat less capable of exploiting the on hand solutions than the exploration of search space. Therefore unlike the deterministic methods, DE has intrinsic negative aspect of skipping the accurate optima. This paper incorporates the memetic search strategy inspired by golden section search [42] which exploits the best solution after each and every iteration in order to generate its neighborhood solutions. To validate the proposed MSDE's performance, experiments are carried out on 11 benchmark as well as real life problems of different complexities and results reflect the superiority of the proposed strategy than the basic DE.

Rest of the paper is systematized as follows: Sect. 2 describes brief overview of the basic DE. Memetic algorithms explained in Sect. 3. Memetic search in DE (MSDE) is proposed and tested in Sect. 4. In Sect. 5, a comprehensive set of experimental results are provided. Finally, in Sect. 6, paper is concluded.

## 2. DIFFERENTIAL EVOLUTION ALGORITHM

Differential evolution is a strategy that optimizes a dilemma by iteratively trying to enhance an individual solution with regard to a specified gauge of excellence. DE algorithm is used for multidimensional real-valued functions but it does not put together the ascent of the problem being optimized, which means DE does not have need of that the optimization problem to be differentiable as is mandatory for traditional optimization methods such as gradient descent and quasi-Newton techniques. DE algorithm optimizes a problem by considering a population of candidate solutions and generating new contestant solutions by combining existing ones according to its straightforward formulae, and then memorizing whichever candidate solution has the superlative score or fitness on the optimization problem at hand. Thus in this manner the optimization problem is treated as a black box that simply makes available a gauge of quality specified a candidate solution and the gradient is for that reason not considered necessary. DE has a number of strategies based on method of selecting the objective vector, number of difference vectors used and the crossover type [8]. Here in this paper DE/rand/1/bin scheme is used where DE stands for differential evolution, 'rand' indicates that the target vector is preferred haphazardly, '1' is for number of differential vectors and 'bin' notation is for binomial crossover. The popularity of Differential Evolution is due to its applicability to a wider class of problems and ease of implementation. Differential Evolution has properties of evolutionary algorithms as well as swarm intelligence. The detailed description of DE is as follows:

In the vein of other population based search algorithms, in DE a population of probable solutions (individuals) searches the solution. In a D-dimensional search space, an individual is represented by a D-dimensional vector (xi1, xi2, . . ., xiD), i = 1, 2, . . ., NP where NP is the population size (number of individuals). In DE, there are three operators: mutation, crossover and selection. Initially, a population is generated randomly with uniform distribution then the mutation, crossover and selection operators are applied to generate a new population. Trial vector generation is a critical step in DE progression. The two operators, mutation and crossover are used to engender the tryout vectors. The selection operator is used to decide on the best tryout vector for the subsequently age bracket. DE operators are explained for a split second in following subsections.

### 2.1 Mutation

A tryout vector is generated by the DE mutation operator for each individual of the in progress population. For generating the tryout vector, an objective vector is mutated with a biased differential. A progeny is fashioned in the crossover operation using the recently generated tryout vector. If G is the index for generation counter, the mutation operator for generating a trial vector $v_i(G)$ from the parent vector $x_i(G)$ is defined as follows:

- Choose a objective vector $x_{i1}(G)$ from the population such that $i$ and $i_1$ are poles apart.
- All over again, haphazardly pick two individuals, $x_{i2}$ and $x_{i3}$, from the population such that $i, i_1, i_2$ and $i_3$ all are distinct to each other.
- After that the objective vector is mutated for calculating the tryout vector in the following manner:

$$v_i(G) = x_{i1}(G) + F * (x_{i2}(G) - x_{i3}(G)) \qquad (1)$$

Here $F \in [0, 1]$ is the mutation scale factor which is used in controlling the strengthening of the differential variation [7].

### 2.2 Crossover

Offspring $x'_i(G)$ is generated using the crossover of parent vector $x_i(G)$ and the tryout vector $u_i(G)$ as follows:

$$x'_{ij}(G) = \begin{cases} v_{ij(G)}, & if\ j \in J \\ x_{ij}(G), & otherwise \end{cases} \qquad (2)$$

Here $J$ is the set of cross over points or the points that will go under perturbation, $x_{ij}(G)$ is the $j^{th}$ element of the vector $x_i(G)$. Diverse methods possibly will be used to settle on the set $J$ in which binomial crossover and exponential crossover are the most commonly used [7]. Here this paper used the binomial crossover. In this crossover, the crossover points are arbitrarily preferred from the set of potential crossover points, {1, 2, . . ., D}, where D is the dimension of problem.

### 2.3 Selection

There are a couple of functions for the selection operator: First it selects the individual for the mutation operation to generate the trial vector and second, it selects the most excellent, between the parent and the offspring based on their fitness value for the next generation. If fitness of parent is superior than the offspring then parent is selected otherwise offspring is selected:

$$x_i(G+1) = \begin{cases} x'_i(G), & if\ f(x'_i(G)) > f(x_i(G)) \\ x_i(G), & otherwise \end{cases} \qquad (3)$$

The above equation makes sure that the population's average fitness does not get worse. The Pseudo-code for Differential Evolutionary strategy is described in Algorithm 1 [7].

---

**Algorithm 1 Differential Evolution Algorithm**

Initialize the control parameters, *F* and *CR* (*F* (scale factor) and *CR* (crossover probability)).

Generate and initialize the population, *P(0)*, of *NP* individuals (*P* is the population vector)

while stopping criteria not meet do

   for each individual, $x_i(G) \in P(G)$ do

     Estimate the fitness, $f(x_i(G))$;

     Generate the trial vector, $v_i(G)$ by using the mutation operator;

     Generate an offspring, $x'_i(G)$, by using the crossover operator;

     if $f(x'_i(G))$ is better than $f(x_i(G))$ then

        Add $x'_i(G)$ to $P(G + 1)$;

     else

        Add $x_i(G)$ to $P(G + 1)$;

     end if

   end for

end while

Memorize the individual with the best fitness as the solution

---

## 3. MEMETIC ALGORITHMS

Memetic algorithms (MA) characterize one of the most up to date mounting fields to do research in evolutionary computation. The name MA is now a day commonly used as coactions of evolutionary or several population-based algorithms with separate individual learning or local development events for problem search. Sometimes MA is moreover specified in the text as Baldwinian evolutionary algorithms (EA), Lamarckian EAs, genetic local search strategy or cultural algorithms. The theory of "Universal Darwinism" was given by Richard Dawkins in 1983[12] to make available a amalgamate structure governing the evolution of any intricate system. In particular, "Universal Darwinism" put forward that development is not restricted to biological systems; i.e., it is not limited to the tapered circumstance of the genes, but it is applicable to almost all multifarious system that show evidence of the principles of inheritance, disparity and assortment, thus satisfying the individuality of an evolving system. For instance, the new science of memetics symbolizes the mind-universe corresponds to genetics in way of life progression that stretches transversely the fields of biology, psychology and cognition, which has fascinated significant concentration in the last two decades. The term "meme" was also introduced and defined by Dawkins in 1976[13] as "the basic unit of cultural transmission, or imitation", and in the English Dictionary of Oxford as "an element of culture that may be considered to be passed on by non-genetic means".

Inspired by both Darwinian philosophy of ordinary evolution and Dawkins' conception of a meme, the term "Memetic Algorithm" (MA) was first given by Moscato in his technical report [14] in 1989 where he observed MA as being very analogous to a outward appearance of population-based hybrid genetic algorithm (GA) together with an individual learning modus operandi talented to perform local refinements. The allegorical analogous, one side occupied by Darwinian evolution and, other side, between memes and domain specific (local search) heuristics are incarcerate within memetic algorithms as a consequence rendering a methodology that balances well between generality and problem specificity. In a more assorted context, memetic algorithms are now used under a variety of names together with Baldwinian Evolutionary Algorithms, Cultural Algorithms, Genetic Local Search, Hybrid Evolutionary Algorithms and Lamarckian Evolutionary Algorithms. In the circumstance of intricate optimization, many different instantiations of memetic algorithms have been reported athwart an extensive range of application domains, in general, converging to premium solutions more efficiently than their unadventurous evolutionary matching part. In broad-spectrum, using the thoughts of memetics surrounded by a computational structure is called "Memetic Computing or Memetic Computation" (MC) [15]16] With MC, the individuality of Universal Darwinism are more appropriately captured. In all viewpoints, MA is a more constrained impression of MC. Exclusively; MA covers vicinity of MC, in meticulous dealing with areas of evolutionary algorithms that get hitched other deterministic enhancement techniques for solving optimization problems. MC lengthens the impression of memes to cover up conceptual entities of knowledge-enhanced measures or representations.

Memetic algorithms are the area under discussion of intense scientific research and have been fruitfully applied to a massive amount of real-world problems. Even though many people make use of techniques intimately associated to memetic algorithms, unconventional names such as hybrid genetic algorithms are also in employment. Moreover, many people name their memetic techniques as genetic algorithms. The prevalent use of this misnomer hampers the measurement of the total amount of applications. MAs are a course group of stochastic global search heuristics in which Evolutionary Algorithms-based approaches are pooled with problem-specific solvers. After that might be implemented as local search heuristics techniques, approximation algorithms or, sometimes, even precise methods. The hybridization is preordained to either speed up the discovery of good solutions, for which evolution alone would take too long to discover, or to attain solutions that would otherwise be out-of-the-way by evolution or a local method alone. As the great preponderance of Memetic Algorithms use heuristic local searches rather than precise methods. It is tacit that the evolutionary search provides for an extensive exploration of the search space while the local search by some means zoom-in on the sink of magnetism of talented solutions Researchers have used memetic algorithms to embark upon many conventional NP problems. To mention some of them: bin packing problem, generalized assignment problem, graph partitioning, max independent set problem, minimal graph coloring, multidimensional knapsack, quadratic assignment problem, set cover problem and travelling salesman problem.

MAs have been demonstrated very triumphant transversely an extensive range of problem domains. More topical applications take account of (but are not limited to) artificial neural network training,[17] pattern recognition,[18] motion planning in robotic,[19] beam orientation,[20] circuit design,[21] electric service restoration,[22] expert systems for medical field,[23] scheduling on single machine,[24] automatic timetabling,[25] manpower scheduling,[26] nurse rostering and function optimization [27] allocation of processor,[28] maintenance scheduling (for instance, of an electric distribution network),[29] multidimensional knapsack problem by E ozcan et al.[30] VLSI design,[31] clustering of gene expression profiles,[32] feature/gene selection or

extraction,[33][34] and multi-class, multi-objective feature selection.[35][36]. M Marinaki et al. [37] proposed an island memetic differential evolution algorithm, for solving the feature subset selection problem while the nearest neighbor categorization method is used for the classification task. CG Uzcátegui et al. [38] developed a memetic differential evolution algorithm to solve the inverse kinematics problem of robot manipulators. S Goudos [39] designed a memetic differential evolution algorithm to design multi-objective approach to sub arrayed linear antenna arrays. Memetic algorithm also hybridized with artificial bee colony algorithm like JC Bansal et al. [40] anticipated a memetic search in artificial bee colony algorithm. S Kumar et al. [41] developed a randomized memetic artificial bee colony algorithm and applied those algorithms to solve various problems.

## 4. MEMETIC SEARCH IN DIFFERENTIAL EVOLUTION (MSDE) ALGORITHM

Exploration of the entire search space and exploitation of the most favorable solution region in close proximity may be unprejudiced by maintaining the diversity in the early hours and later on iterations of any random number based search algorithm. It is unambiguous from the Eqs. (1) and (2) that exploration of the search space in DE algorithm controlled by two parameters (*CR* and *F*). In case of DE algorithm, exploration and exploitation of the search space depend on the value of *CR* and *F* that is to say if value of *CR* and *F* are high then exploration will be high and exploitation will be high if value of *CR* and *F* are low. This paper introduced a new search strategy in order to balance the process of exploration and exploitation of the search space. The new search strategy is inspired from the memetic search in Artificial Bee Colony (MeABC) algorithm [40]. The MeABC algorithm developed by J. C. Bansal et al [40] aggravated by Golden Section Search (GSS) [42]. In MeABC only the best individual of the in progress population updates themselves in its concurrence. GSS processes the interval [a = −1.2, b = 1.2] and initiates two intermediate points ($f_1$, $f_2$) with help of golden ratio ($\Psi = 0.618$). The GSS process summarized in algorithm 2 as follow:

---

**Algorithm 2: Golden Section Search**
Repeat while termination criteria meet
Compute $f_1 = b - (b-a) \times \psi$, and $f_2 = a + (b-a) \times \psi$,
Calculate $f(f_1)$ and $f(f_2)$
If $f(f_1) < f(f_2)$ then
$b = f_2$ and the solution lies in the range [a, b]
else
$a = f_1$ and the solution lies in the range [a, b]
End of if
End of while

---

The proposed memetic search strategy modifies the mutation operator of DE algorithm and modifies the Eqs. (1) as follow:

$v_i(G) = x_{i1}(G) + F * (x_{i2}(G) - x_{i3}(G)) + f_j * (x_{i2}(G) - x_{i3}(G))$

Here $f_j$ is decided by algorithm 2.

The proposed changes in original DE algorithm control the search process adaptively and provide more chances to explore the large search space and exploit the better solution in more efficient way. This change in DE tries to balance intensification and diversification of search space. The detailed Memetic search in DE (MSDE) outlined in algorithm 3 as follow:

---

**Algorithm 3 Memetic Search in Differential Evolution (MSDE) Algorithm**

Initialize the control parameters, *F* and *CR* (*F* (scale factor) and *CR* (crossover probability)).

Generate and initialize the population, *P(0)*, of *NP* individuals (*P* is the population vector)

while stopping criteria not meet do
    for each individual, $x_i(G) \in P(G)$ do
        Estimate the fitness, $f(x_i(G))$;
        Generate the trial vector, $v_i(G)$ by using the mutation operator and incorporate algorithm 2 as per the following equation;
$$v_i(G) = x_{i1}(G) + F * (x_{i2}(G) - x_{i3}(G)) + f_j * (x_{i2}(G) - x_{i3}(G))$$
        Generate an offspring, $x'_i(G)$, by using the crossover operator;
        if $f(x'_i(G))$ is better than $f(x_i(G))$ then
            Add $x'_i(G)$ to $P(G + 1)$;
        else
            Add $x_i(G)$ to $P(G + 1)$;
        end if
    end for
end while
Memorize the individual with the best fitness as the solution

---

## 5. EXPERIMENTAL RESULTS
### 5.1 Test problems under consideration

Differential Evolution algorithm with improvement in search phase applied to the eight benchmark functions for whether it gives better result or not at different probability and also applied for three real world problems. Benchmark functions taken in this paper are of different characteristics like uni-model or multi-model and separable or non-separable and of different dimensions. In order to analyze the performance of MSDE, it is applied to global optimization problems ($f_1$ to $f_8$) listed in Table I. Test problems $f_1$–$f_8$ are taken from [8],[44]. Three real world problems (Pressure Vessel Design, Lennard Jones and Parameter estimation for frequency modulated sound wave) are described as follow:

**Pressure Vessel Design ($f_9$):** The problem of minimizing total cost of the material, forming and welding of a cylindrical vessel [43]. In case of pressure vessel design generally four design variables are considered: shell thickness ($x_1$), spherical head thickness ($x_2$), radius of cylindrical shell ($x_3$) and shell length ($x_4$). Simple mathematical representation of this problem is as follow:

$$f_9 = 0.6224 x_1 x_3 x_4 + 1.7781 x_2 x_3^2 + 3.1611 x_1^2 x_4 + 19.84 x_1^2 x_3$$

Subject to

$$g_1(x) = 0.0193 x_3 - x_1, g_2(x) = 0.00954 x_3 - x_2,$$

$$g_3(x) = 750 * 1728 - \pi x_3^2 (x_4 + \frac{4}{3} x_3)$$

The search boundaries for the variables are

$1.125 \leq x_1 \leq 12.5$, $0.625 \leq x_2 \leq 12.5$,

$1.0*10^{-8} \leq x_3 \leq 240$ and $1.0*10^{-8} \leq x_4 \leq 240$.

The best ever identified global optimum solution is $f(1.125, 0.625, 55.8592, 57.7315) = 7197.729$ [43]. The tolerable error for considered problem is $1.0E-05$.

**Table 1. Test Problems**

| Test Problem | Objective Function | Search Range | Optimum Value | D | Acceptable Error |
|---|---|---|---|---|---|
| Step Function | $f_1(x) = \sum_{i=1}^{D}(\lfloor x_i + 0.5 \rfloor)^2$ | [-100, 100] | $f(-0.5 \leq x \leq 0.5)=0$ | 30 | 1.0E-05 |
| Colville function | $f_2(x) = 100(x_2 - x_1^2)^2 + (1-x_1)^2 + 90(x_4 - x_3^2)^2 + (1-x_3)^2 + 10.1[(x_2-1)^2 + (x_4-1)^2] + 19.8(x_2-1)(x_4-1)$ | [-10, 10] | $f(1) = 0$ | 4 | 1.0E-05 |
| Kowalik function | $f_3(x) = \sum_{i=1}^{11}(a_i - \frac{x_1(b_i^2 + b_i x_2)}{b_i^2 + b_i x_3 + x_4})^2$ | [-5, 5] | $f(0.1928, 0.1908, 0.1231, 0.1357) = 3.07E-04$ | 4 | 1.0E-05 |
| Shifted Rosenbrock | $f_4(x) = \sum_{i=1}^{D-1}(100(z_i^2 - z_{i+1})^2 + (z_i - 1)^2 + f_{bias}$, $z = x - o + 1, x = [x_1, x_2, ... x_D], o = [o_1, o_2, ...... o_D]$ | [-100, 100] | $f(o)=f_{bias}=390$ | 10 | 1.0E-01 |
| Six-hump camel back | $f_5(x) = (4 - 2.1x_1^2 + \frac{1}{3}x_1^4)x_1^2 + x_1 x_2 + (-4 + 4x_2^2)x_2^2$ | [-5, 5] | $f(-0.0898, 0.7126) = -1.0316$ | 2 | 1.0E-05 |
| Hosaki Problem | $f_6(x) = (1 - 8x_1 + 7x_1^2 - \frac{7}{3}x_1^3 + \frac{1}{4}x_1^4)x_2^2 \exp(-x_2)$ | $x_1 \in [0,5]$, $x_2 \in [0,6]$ | -2.3458 | 2 | 1.0E-06 |
| Meyer and Roth Problem | $f_7(x) = \sum_{i=1}^{5}(\frac{x_1 x_3 t_i}{1 + x_1 t_i + x_2 v_i} - y_i)^2$ | [-10, 10] | $f(3.13, 15.16, 0.78) = 0.4E-04$ | 3 | 1.0E-03 |
| Shubert | $f_8(x) = -\sum_{t=1}^{5} i\cos((i+1)x_1 + 1) \sum_{i=1}^{5} i\cos((i+1)x_2 + 1)$ | [-10, 10] | $f(7.0835, 4.8580) = -186.7309$ | 2 | 1.0E-05 |

**Parameter Estimation for Frequency-Modulated (FM) sound wave ($f_{11}$):** Frequency-Modulated (FM) sound wave amalgamation has a significant role in several modern music systems. The parameter optimization of an FM synthesizer is an optimization problem with six dimension where the vector to be optimized is X = {$a_1$, $w_1$, $a_2$, $w_2$, $a_3$, $w_3$} of the sound wave given in equation (4). The problem is to generate a sound (1) similar to target (2). This problem is a highly complex multimodal one having strong epistasis, with minimum value $f(X) = 0$. This problem has been tackled using Artificial bee colony algorithms (ABC) in [43]. The expressions for the estimated sound and the target sound waves are given as:

$$y(t) = a_1 \sin(w_1 t\theta + a_2 \sin(w_2 t\theta + a_3 \sin(w_3 t\theta))) \quad (4)$$
$$y_0(t) = (1.0)\sin((5.0)t\theta - (1.5)\sin((4.8)t\theta + (2.0)\sin((4.9)t\theta))) \quad (5)$$

Respectively where $\theta = 2\pi/100$ and the parameters are defined in the range [-6.4, 6.35]. The fitness function is the summation of square errors between the estimated wave (1) and the target wave (2) as follows:

$$f_{11}(x) = \sum_{i=0}^{100}(y(t) - y_0(t))^2$$

Acceptable error for this problem is 1.0E-05, i.e. an algorithm is considered successful if it finds the error less than acceptable error in a given number of generations.

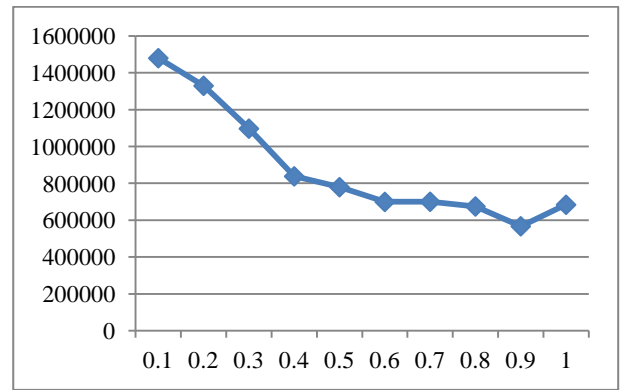

**Fig 1: Effect of CR on AFE**

## 5.2 Experimental Setup
To prove the efficiency of MSDE, it is compared with original DE algorithm over well thought-out seventeen problems, following experimental setting is adopted:
- Population Size NP = 50
- The Scale factor F= 0.5
- Limit = D*NP/2.
- Number of Run =100
- Sopping criteria is either reached the corresponding acceptable error or maximum function evaluation (which is set as 200000).
- Crossover probability CR = 0.9, in order to identify the effect of the parameter CR on the performance

of MSDE, its effect at different CR in range [0.1, 1] is experimented as shown in figure 1. It can be easily observed from the graph that best value of CR is 0.9 for considered as problems.

- The mean function values (MFV), standard deviation (SD), mean error (ME), average function evaluation (AFE) and success rate (SR) of considered problem have been recorded.
- Experimental setting for DE is same as MSDE.

### 5.3 Result Comparison

Mathematical results of MSDE with experimental setting as per section 5.2 are discussed in Table 2. Table 2 show the relationship of results based on mean function value (MFV), standard deviation (SD), mean error (ME), average function evaluations (AFE) and success rate (SR). Table 2 shows that a good number of the times MSDE outperforms in terms of efficiency (with less number of function evaluations) and reliability as compare to other considered algorithms. The proposed algorithm all the time improves AFE. It is due to balancing between exploration and exploitation of search space. Table 3 contains summary of table 2 outcomes. In Table 3, '+' indicates that the MSDE is better than the well thought-out algorithms and '-' indicates that the algorithm is not better or the divergence is very small. The last row of Table 3, establishes the superiority of MSDE over DE.

**Table 2. Comparison of the results of test problems**

| Test Problem | Algorithm\Measure | MFV | SD | ME | AFE | SR |
|---|---|---|---|---|---|---|
| $f_1$ | DE | 1.20E-01 | 4.75E-01 | 1.20E-01 | 31942.5 | 91 |
| | MSDE | 1.00E-02 | 9.95E-02 | 2.00E-02 | 21319 | 99 |
| $f_2$ | DE | 1.47E-01 | 6.46E-01 | 1.47E-01 | 26918 | 89 |
| | MSDE | 4.57E-04 | 5.12E-04 | 7.50E-04 | 31196 | 95 |
| $f_3$ | DE | 5.72E-04 | 3.29E-04 | 2.65E-04 | 61900 | 71 |
| | MSDE | 5.53E-04 | 3.55E-04 | 2.51E-04 | 74294 | 82 |
| $f_4$ | DE | 3.92E+02 | 2.09E+00 | 2.13E+00 | 194913.5 | 3 |
| | MSDE | 3.91E+02 | 2.49E+00 | 8.65E-01 | 197758 | 7 |
| $f_5$ | DE | -1.03E+00 | 1.42E-05 | 1.79E-05 | 112618 | 44 |
| | MSDE | -1.03E+00 | 4.58E-06 | 1.12E-05 | 57491.5 | 72 |
| $f_6$ | DE | -2.35E+00 | 5.96E-06 | 5.26E-06 | 10858 | 95 |
| | MSDE | -2.35E+00 | 2.53E-06 | 5.19E-06 | 1261 | 100 |
| $f_7$ | DE | 1.91E-03 | 1.64E-05 | 1.95E-03 | 3744 | 99 |
| | MSDE | 1.90E-03 | 2.33E-06 | 1.95E-03 | 4621 | 99 |
| $f_8$ | DE | -1.87E+02 | 5.37E-06 | 4.59E-06 | 8122 | 100 |
| | MSDE | -1.87E+02 | 2.53E-06 | 4.66E-06 | 7375.5 | 100 |
| $f_9$ | DE | 7.20E+03 | 3.38E-05 | 2.43E-05 | 65912.5 | 71 |
| | MSDE | 7.20E+03 | 1.51E-05 | 7.88E-06 | 19772 | 96 |
| $f_{10}$ | DE | -9.10E+00 | 1.45E-05 | 8.13E-05 | 72866 | 100 |
| | MSDE | -9.10E+00 | 1.65E-05 | 7.71E-05 | 69196.5 | 100 |
| $f_{11}$ | DE | 5.83E+00 | 6.33E+00 | 5.83E+00 | 113179 | 48 |
| | MSDE | 4.83E+00 | 6.49E+00 | 4.84E+00 | 83060 | 63 |

## 6. CONCLUSION

This paper, modify the search process in original DE by introducing customized Golden Section Search process. Newly introduced line of attack added in mutation process. Proposed algorithm modifies step size with the help of GSS process and update target vector. Further, the modified strategy is applied to solve 8 well-known standard benchmark functions and three real world problems. With the help of experiments over test problems and real world problems, it is shown that the insertion of the proposed strategy in the original DE algorithm improves the steadfastness, efficiency and accuracy as compare to its original version. Table 2 and 3 shows that the proposed MSDE is able to solve almost all the considered problems with fewer efforts. Numerical results show that the improved algorithm is superior to original DE algorithm. Proposed algorithm has the ability to get out of a local minimum and has higher rate of convergence. It can be resourcefully applied for separable, multivariable, multimodal function optimization. The proposed strategy also improves results for three real world problems: Pressure Vessel Design, Lennard Jones and Parameter estimation for frequency modulated sound wave.

**Table 3. Summary of table II outcome**

| Test Problem | MSDE vs. DE |
|---|---|
| $f_1$ | + |
| $f_2$ | + |
| $f_3$ | + |
| $f_4$ | + |
| $f_5$ | + |
| $f_6$ | + |
| $f_7$ | - |
| $f_8$ | + |
| $f_9$ | + |
| $f_{10}$ | + |
| $f_{11}$ | + |
| Total Number of + sign | 10 |